\documentclass{llncs}

\usepackage{url}
\usepackage{comment}

\usepackage{epsfig}
\usepackage{mdwlist}

\usepackage{rotating}

\usepackage{graphicx}
\usepackage{amssymb}
\usepackage{mathtools}

\usepackage{tikz}

\usepackage{colortbl,hhline}
\usepackage{multirow}

\usetikzlibrary{plotmarks,shapes,arrows,backgrounds,fit,shadows,positioning}

\definecolor{myred}{rgb}{0.8,0,0.1} 
\definecolor{mygreen}{rgb}{0.0,0.7,0.05} 
\definecolor{myblue}{rgb}{0.2,0.1,0.7} 
\definecolor{myorange}{rgb}{0.8,0.5,0.03} 
\definecolor{myviolet}{rgb}{0.7,0.1,0.7} 

\definecolor{grayx}{rgb}{0.9,0.9,0.9}

\def\ONT{\mathcal{O}}
\def\ALI{\mathcal{A}}
\def\REF{\mathcal{R}}

\newcommand{\tick}{\ensuremath{\surd}}
\newcommand{\ignore}[1]{}
\makeatletter
\def\@ydefthm#1#2[#3]{%
    \trivlist
    \item[%
        \hskip 10\p@
        \hskip \labelsep
        {\it #2%
         \ \rm \csname the#1\endcsname
         \ [#3].
        }]%
    \ignorespaces
}

\newcommand{\alignset}{
\tikzstyle{onto}=[shape=circle,minimum size=.8cm,fill=yellow] 

\tikzstyle{align}=[shape=circle,minimum size=.8cm,fill=green] 

\tikzstyle{med}=[shape=rectangle,rounded corners,minimum size=.8cm,fill=orange!20]
\tikzstyle{axioms}=[shape=rectangle,rounded corners,minimum size=.8cm,fill=orange!20]

\tikzstyle{peer}=[ellipse,minimum size=.8cm,fill=purple!20]
\tikzstyle{agent}=[ellipse,minimum size=.8cm,fill=purple!20]
\tikzstyle{ws}=[ellipse,minimum size=.8cm,fill=purple!20] 

\tikzstyle{message}=[shape=rectangle,minimum size=.8cm,fill=orange!20]
\tikzstyle{query}=[shape=rectangle,minimum size=.8cm,fill=yellow!20!white] 
\tikzstyle{answer}=[shape=rectangle,minimum size=.8cm,fill=blue!20] 

\tikzstyle{arrow}=[>=latex,->]
\tikzstyle{arrowd}=[>=latex,<->]
\tikzstyle{app}=[>=latex,->,thin]
\tikzstyle{uses}=[>=stealth,->>,thin]
\tikzstyle{interp}=[>=latex,->,thin]

\tikzstyle{aggr}=[shape=isosceles triangle,minimum size=1cm,fill=orange!20]
\tikzstyle{disp}=[shape=isosceles triangle,rotate=180,minimum size=1cm,fill=orange!20]

\tikzstyle{matchers}=[shape=rectangle, rounded corners, minimum size=.8cm,draw, double copy shadow={shadow xshift=1ex,shadow yshift=1ex},fill=orange!20,thick] 
\tikzstyle{arrowDD}=[>=latex,->,densely dashed]
\tikzstyle{database}=[shape=cylinder, shape border rotate=90,draw, minimum height=1.2cm, minimum width=1.0cm,aspect=0.5]

}

\newcommand{\algoset}{
\alignset
\tikzstyle{proc}=[shape=rectangle, text width=1.5cm, text badly centered, minimum size=.8cm,draw] 
\tikzstyle{matrix}=[shape=circle,minimum size=.5cm,fill=orange]
\tikzstyle{afilter}=[shape=or gate US,fill=blue!20]
}

\begin{document}

\title{Evaluating Ontology Matching Systems on Large, Multilingual and Real-world Test Cases}


\author{C. Meilicke\inst{1}, 
O. \v{S}v\'{a}b-Zamazal\inst{3}, 
C. Trojahn\inst{2},
E. Jim\'{e}nez-Ruiz\inst{4},\\ 
J.L. Aguirre\inst{2},
H. Stuckenschmidt\inst{1},
B. Cuenca Grau\inst{4}}

\institute{University of Mannheim
\and INRIA \& LIG, Grenoble
\and University of Economics, Prague
\and University of Oxford, UK}
\maketitle

\begin{abstract}
In the field of ontology matching, the most systematic evaluation of matching systems is established by the Ontology Alignment Evaluation Initiative (OAEI), 
which is an annual campaign for evaluating ontology matching systems organized by different groups of researchers. In this paper, 
we report on the results of an intermediary OAEI campaign called OAEI 2011.5. The evaluations of this campaign are divided in five tracks. 
Three of these tracks are new or have been improved compared to previous OAEI campaigns.
Overall, we evaluated 18 matching systems. 
We discuss lessons learned, in terms of scalability, multilingual issues and the ability do deal with real world cases from different domains.
\end{abstract}

\section{Introduction}
\label{sec:intro}
The development in the area of semantic technologies has been enabled by the standardization of knowledge representation languages on the web, 
in particular RDF and OWL. Based on these languages, many tools have been developed to perform various tasks on the semantic web, 
such as searching, querying, integrating and reasoning about semi-structured information. However, a crucial step in their large scale 
adoption in real world applications is the ability to determine the quality of a system in terms of its expected performance on realistic data. 
Semantic technologies, even though they support a similar functionality, are
often not evaluated against the same data sets or the measured results are reproducible with significant effort only. 
Hence, the challenges on semantic technologies evaluation involves (a) the evaluation of technologies on the basis of test cases that allow conclusions relevant for real 
world applications and (b) the automatism and reproducibility of the evaluation process and its results. 


Regarding the first point, in the field of ontology matching, systematic evaluations are established by the  
Ontology Alignment Evaluation Initiative (OAEI)~\cite{euzenat11sixyears}.
It is an annual evaluation campaign, carried out since 2004, that offers datasets, from different domains, organized by different groups of researchers. Recently, two new datasets have been proposed~\cite{multifarm2012,trackBioMed2011} that put a special focus on scalability and multilingual coverage. These are important aspects, due to recent initiatives such as Open Linked Data, where a large amount of multilingual data has to be processed. The aim of this paper is to report on the evaluation results of an intermediary OAEI campaign, called OAEI 2011.5, that includes these two datasets.
With respect to the second point, the SEALS project
(\url{http://about.seals-project.eu/}) has focused on establishing automatic and systematic evaluation methods for semantic technologies by providing, 
in particular, a software infrastructure for automatically executing evaluations.
This infrastructure involves a controlled execution environment where evaluation organizers can run a set of tools on the same data set. 
Tools, test data and results in the context of an evaluation campaign are stored in the SEALS repositories. The OAEI 2011.5 campaign is executed on top of this infrastructure. This allows to reproduce all evaluation results that are reported within this paper. 


First, we describe ontology matching and the evaluation of ontology matching
systems in~\S\ref{sec:background}. In~\S\ref{sec:methodology} we continue with
a description of the experimental setting that we applied to OAEI 2011.5. We
present the results of our evaluation experiments for each dataset on its own
in \S\ref{sec:results}.1-\S\ref{sec:results}.5. Finally, we summarize the most
important lessons learned in~\S\ref{sec:lessons}.

\section{Ontology matching evaluation}
\label{sec:background}
\begin{figure}[t]
\begin{center}
\begin{tikzpicture}[scale=.75,cap=round]
\algoset;

\draw (-1,1.7) node[style=onto](o){$\ONT_1$};
\draw (-1,0.3) node[style=onto](op){$\ONT_2$};

\draw[fill=white] (2,1) node[style=proc,fill=white](m){matching};
\draw [style=arrow] (o.east) -- (m.west);
\draw [style=arrow] (op.east) -- (m.west);

\draw (2,2.2) node(p){$parameters$};
\draw [style=arrow] (p.south) -- (m.north);

\draw (2,-0.2) node(r){$resources$};
\draw [style=arrow] (r.north) -- (m.south);

\draw (4,1) node[style=align](Ap){$\ALI$};
\draw [style=arrow] (m.east) -- (Ap.west);

\draw (4,2.4) node[style=align](R){$\REF$};

\draw (7,1.7) node[style=proc](ev){evaluator};
\draw [style=arrow] (R.east) -- (ev.west);
\draw [style=arrow] (Ap.east) -- (ev.west);

\draw (9,1.7) node[circle](ms){$m$};
\draw [style=arrow] (ev.east) -- (ms.west);

\end{tikzpicture} 
\end{center}
\caption{Ontology matching process and evaluation (from \cite{euzenat2007b}).}
\label{fig:figmatch}
\end{figure}
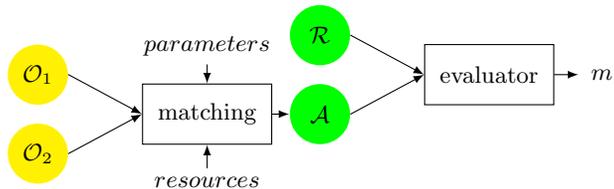

There have been different formalizations of the matching process~\cite{bernstein2000b,lenzerini2002a}.
We follow the framework presented in~\cite{euzenat2007b}
(see Figure~\ref{fig:figmatch}). According to this framework, ontology
matching systems generate alignments that are sets of correspondences. Given two ontologies $\ONT_1$ and $\ONT_2$, an example for a correspondence is the statement that \texttt{SubjectArea} in $\ONT_1$ is the same as a \texttt{Topic} in $\ONT_2$ or that \texttt{ExternalReviewer} in $\ONT_1$ is a subclass of \texttt{Reviewer} in $\ONT_2$. 
In this example, one of the correspondences expresses an equivalence, while the other one expresses a subsumption relation. 
The core elements of a correspondence are an entity from $\ONT_1$, an entity from $\ONT_2$, and a relation that is supposed to 
hold between them. The matched entities can be classes, properties or instances. In our experiments we are only concerned with matching classes and properties via equivalence.

A minimal data set for evaluating ontology matching systems consists of two
ontologies $\ONT_1$ and $\ONT_2$ and an alignment $\REF$ that is used as a gold
standard. The quality of an alignment $\ALI$ is measured in terms of its
compliance (precision and recall) against the reference alignment $\REF$. 
Precision is defined as $| \ALI \cap \REF | / |\ALI|$, while recall is defined
as $| \ALI \cap \REF | / |\REF|$. The F-measure combines precision and recall
and is usually represented as their harmonic mean.
For most of our experiments we present aggregated values for these three
measures.

In addition to these compliance based measures, it is also important to measure
the runtime of a matching process and to understand what factors have an impact
on runtime and alignment quality (size of the ontologies, available
resources in terms of computational power and additional background
knowledge). We have, for example, conducted specific experiments to see whether
a matching system can exploit a multi-core architecture to speed up the matching
process. Another criteria is the coherence of the generated alignment as
defined in~\cite{meilicke08imeasure}. The coherence of an alignment $\ALI$ is
commonly measured with respect to the number of unsatisfiable classes obtained
when reasoning with the input ontologies $\ONT_1$ and $\ONT_2$ together with
$\ALI$. The coherence of an alignment
is very important and determines whether it can be used for certain
types of applications (e.g., query processing and data migration) that require
coherent alignments.



\section{Experimental setting}
\label{sec:methodology}

Before describing data sets
and tools,
we give a brief overview on the overall procedure of OAEI 2011.5. The whole campaign can be divided in three phases:
\begin{description}
	\item[Preparatory] Participants wrap their tools against a predefined interface. Thus,  evaluations can be executed locally by using a provided client software. This allows to check whether their tool works correctly with the data sets.
   \item[Execution] Final tool versions are uploaded by participants and the organizers run the evaluation using SEALS infrastructure with both blind and published datasets. Generated results are stored in the SEALS repository.
   \item[Evaluation]  Stored results are analyzed, aggregated, visualized and
   published. An extended report on all OAEI 2011.5 results can also be found at
   \url{http://oaei.ontologymatching.org/2011.5/results/index.html}
\end{description}
We have run evaluation experiments divided in five different tracks. The tracks MultiFarm and Large BioMed appear for the first time in an OAEI campaign. 
\begin{description}
	\item[Benchmarks] For this track, the focus of this campaign was on
	scalability; to that extent, we considered four ``seed ontologies'' from
	different domains and with different sizes. Two of them are completely new (\texttt{jerm} and \texttt{provenance}), \texttt{biblio} and \texttt{finance} were already considered in OAEI 2011. All data sets were created artificially by a test generator.
	\item[Conference] The Conference track uses a collection of ontologies from the domain of conference organization~\cite{OntoFarm2005}. 
        The ontologies have been created manually by different people and are of moderate size (between 14 and 140 concepts and properties). Reference alignments for a subset of 7 ontologies have been created manually and used since 2008 in OAEI campaigns.
	\item[Anatomy] The anatomy track is about matching the Adult Mouse Anatomy
	(2744 classes) and parts of the NCI Thesaurus (3304 classes) describing the
	human anatomy. The reference alignment, which contains approximately 1000
	correspondences has been created by domain experts~\cite{zhang2004a}. Aside
	from some small modifications, the data set has been used for OAEI since 2007.
	\item[MultiFarm] This track is based on translating the OntoFarm collection to 9 different languages 
         (English, Chinese, Czech, Dutch, French, German, Portuguese, Russian, and Spanish). As a results of this, 
        the track offers challenging test cases for 36 language pairs (further details can be found in~\cite{multifarm2012}).
	\item[Large BioMed] This track aims at finding alignments between
	large and semantically rich biomedical ontologies such as FMA, SNOMED CT,
	and NCI \cite{trackBioMed2011}. In the OAEI 2011.5 we have evaluated the
	FMA-NCI matching problem using two reference alignments based on the UMLS
	Metathesaurus \cite{UMLS2004}.
\end{description}



\setlength{\tabcolsep}{2.75pt}
\begin{table}[t]
\caption{Participation in OAEI 2011 and OAEI 2011.5 tracks B=Benchmarks, C=Conference, M=MultiFarm, A=Anatomy, and L=Large BioMed.}
\label{table:participants}
\vspace{-5pt}
\scriptsize
\begin{center}
\begin{tabular}{|l||c|c||c|c|c|c|c||l|}
\hline
\textbf{System}   & \textbf{2011}  &  \textbf{2011.5} & \textbf{B} & \textbf{C} & \textbf{M} & \textbf{A} & \textbf{L} & \textbf{State, University} \\
\hline
\hline
AgrMaker \cite{agremaker2009}  & \tick &           &  & \tick &  & \tick &  & US, University of
Illinois at Chicago
\\
\hline
Aroma \cite{aroma2007}    & \tick &                 & \tick & \tick &  & \tick & \tick & France, INRIA Grenoble
Rh\^one-Alpes \\
\hline
AUTOMSv2~\cite{automs2012}  &            & \tick     & \tick & \tick & \tick &  &  & Finland, VTT Technical Research Centre  \\
\hline
CIDER \cite{cider2011}    & \tick &                  &  & \tick & \tick &   &  & Spain, Universidad
Polit\'ecnica de Madrid
\\
\hline
CODI~\cite{codibasis2010} & \tick &  \tick       & \tick & \tick & \tick & \tick &  & Germany, Universit\"at
Mannheim \\
\hline
CSA~\cite{csa2011}       & \tick &                                 &  & \tick & \tick & \tick & \tick &  Vietnam,  University of Ho Chi Minh City\\
\hline
GOMMA~\cite{gomma2011}    &             & \tick       & \tick & \tick &  & \tick & \tick &   Germany, Universit\"at
Leipzig\\
\hline
Hertuda   &             & \tick                       & \tick & \tick &  &  &  &   Germany, TU Darmstadt \\
\hline
LDOA      & \tick       &                            &  & \tick &  &  &  &   Tunisia, Tunis-El Manar University \\
\hline
Lily~\cite{lily2011}     & \tick       &                             & \tick & \tick &  & \tick &  &   China, Southeast University \\
\hline
LogMap \cite{logmap_ecai2012}   & \tick &  \tick & \tick & \tick & \tick & \tick & \tick &   UK, University of Oxford \\
\hline
MaasMtch~\cite{maasmatch2011}  & \tick &       \tick  & \tick & \tick & \tick & \tick & \tick &   Netherlands, Maastricht University \\
\hline
MapEVO~\cite{mappsomapevo2011}    & \tick        & \tick  & \tick & \tick & \tick & \tick &  &   Germany, Forschungszentrum Informatik \\
\hline
MapPSO~\cite{mappsomapevo2011}    & \tick        & \tick  & \tick & \tick & \tick & \tick &  &   Germany, Forschungszentrum Informatik \\
\hline
MapSSS~\cite{mapsss2011}    & \tick & \tick         & \tick & \tick & \tick & \tick & \tick &   US, Wright State University  \\
\hline
Optima~\cite{optima2011}    & \tick &              &  & \tick &  &  &  &   US, University of Georgia  \\
\hline
WeSeEMtch &       &       \tick  & \tick & \tick & \tick &  &  &  Germany, TU Darmstadt \\
\hline
YAM++~\cite{yam2011}     & \tick &      \tick  & \tick & \tick &  &  &  &   France, LIRMM \\
\hline
\end{tabular}
\end{center}
\vspace{-15pt}
\end{table}

Table~\ref{table:participants} lists the submitted systems to the SEALS platform
in the OAEI 2011 and 2011.5 campaigns. Note that we have also evaluated
participants of OAEI 2011, always using the most up-to-date version.
As also shown in Table~\ref{table:participants}, not all tools could be
evaluated on all tracks. This is related to problems in processing a certain dataset,
memory exceptions or timeouts. We refer
the reader to the OAEI 2011.5 web page for details that we omit due to the lack of
space.


Note that we have evaluated GOMMA with two different
configurations in Anatomy and Large BioMed tracks. GOMMA$_{\text{bk}}$ uses
specialised background knowledge, while GOMMA$_{\text{nobk}}$ has this feature
deactivated. Furthermore, AgrMaker is
also configured to use specialised background knowledge in Anatomy (referred as
AgrMaker$_{\text{bk}}$).

In addition, we implemented two simple matching algorithms. As
\textit{Baseline-E} we refer to a matcher based on string equality
disregarding capitalization. 
\textit{LogMapLt} is a string
matcher that exploits the creation of an inverted file, a type of index that is
commonly used in information retrieval, to efficiently compute
correspondences.\footnote{See lexical indexation in \cite{logmap_iswc11}.}
In general, recall increases from Baseline-E to LogMapLt, while precision
decreases. 
Note that in many cases it is not easy to top these baselines in terms of
F-measure.

%

\section{Evaluation results and discussion}
\label{sec:results}


\subsection{Benchmarks track}
\label{sec:track-benchmarks}
We considered four ``seed ontologies'' from different domains and with
different sizes.
For each seed ontology, 94 tests were automatically generated. 
Table~\ref{table:benchmarkFMeasure} presents 
the average results for
each benchmark, along with the overall average; values are given
in the format \emph{F-measure (precision$\mid$recall)}. Systems are first
ordered according to the number of benchmarks 
for which an output was provided,
then by the
highest general average. The last column of the table shows the number of
benchmarks for which the matchers generated results and performed at least as
good as the LogMapLt baseline. For example, Aroma passed all benchmarks, and
topped the results of LogMapLt in 3 of them.



\setlength{\tabcolsep}{2pt}
\begin{table}[t]
\caption{Results for benchmark; n/a: not able to run the test,
u/r: uncompleted result.}
\label{table:benchmarkFMeasure}
\vspace{-5pt}
\begin{center}
\scriptsize
\begin{tabular}{|l||c|c|c|c||c||c|}
\hline
\textbf{System} &  \textbf{biblio} &  \textbf{jerm} &  \textbf{provenance} &  \textbf{finance} &  \textbf{avg.} & \textbf{\#} \\
\hline
MapSSS & 0.86 \tiny{(0.99$\mid$0.75)} & 0.76 \tiny{(0.98$\mid$0.63)} & 0.75 \tiny{(0.98$\mid$0.61)} & 0.83 \tiny{(0.99$\mid$0.71)} & 0.80 \tiny{(0.99$\mid$0.68)} & 4/4 \\
\hline
Aroma & 0.76 \tiny{(0.97$\mid$0.63)} & 0.96 \tiny{(0.99$\mid$0.93)} & 0.6 \tiny{(0.78$\mid$0.49)} & 0.7 \tiny{(0.90$\mid$0.57)} & 0.76 \tiny{(0.91$\mid$0.66)} & 3/4 \\
\hline
WeSeE & 0.67 \tiny{(0.89$\mid$0.53)} & 0.68 \tiny{(0.99$\mid$0.51)} & 0.64 \tiny{(0.97$\mid$0.48)} & 0.69 \tiny{(0.96$\mid$0.54)} & 0.67 \tiny{(0.95$\mid$0.52)} & 3/4 \\
\hline
\rowcolor[gray]{.90}
 LogMapLt & 0.58 \tiny{(0.70$\mid$0.50)} & 0.67 \tiny{(0.98$\mid$0.51)} & 0.66 \tiny{(0.99$\mid$0.50)} & 0.66 \tiny{(0.90$\mid$0.52)} & 0.64 \tiny{(0.89$\mid$0.51)} & -- \\
\hline
Hertuda & 0.67 \tiny{(1.00$\mid$0.50)} & 0.66 \tiny{(0.96$\mid$0.50)} & 0.54 \tiny{(0.59$\mid$0.50)} & 0.6 \tiny{(0.75$\mid$0.50)} & 0.62 \tiny{(0.83$\mid$0.50)} & 2/4 \\
\hline
LogMap & 0.48 \tiny{(0.69$\mid$0.37)} & 0.66 \tiny{(1.00$\mid$0.50)} & 0.66 \tiny{(1.00$\mid$0.49)} & 0.6 \tiny{(0.96$\mid$0.43)} & 0.60 \tiny{(0.91$\mid$0.45)} & 2/4 \\
\hline
GOMMA & 0.67 \tiny{(0.79$\mid$0.58)} & 0.67 \tiny{(0.97$\mid$0.51)} & 0.22 \tiny{(0.15$\mid$0.55)} & 0.66 \tiny{(0.84$\mid$0.55)} & 0.56 \tiny{(0.69$\mid$0.55)} & 3/4 \\
\hline
MaasMtch & 0.5 \tiny{(0.49$\mid$0.52)} & 0.52 \tiny{(0.52$\mid$0.52)} & 0.5 \tiny{(0.50$\mid$0.50)} & 0.52 \tiny{(0.52$\mid$0.52)} & 0.51 \tiny{(0.51$\mid$0.52)} & 0/4 \\
\hline
MapPSO & 0.2 \tiny{(0.58$\mid$0.12)} & 0.05 \tiny{(0.06$\mid$0.05)} & 0.07 \tiny{(0.08$\mid$0.05)} & 0.16 \tiny{(0.28$\mid$0.11)} & 0.12 \tiny{(0.25$\mid$0.08)} & 0/4 \\
\hline
MapEVO & 0.37 \tiny{(0.43$\mid$0.33)} & 0.04 \tiny{(0.06$\mid$0.03)} & 0.01 \tiny{(0.02$\mid$0.01)} & 0.02 \tiny{(0.04$\mid$0.01)} & 0.11 \tiny{(0.14$\mid$0.10)} & 0/4 \\
\hline
Lily & 0.75 \tiny{(0.95$\mid$0.62)} & 0.71 \tiny{(0.93$\mid$0.58)} & 0.68 \tiny{(0.92$\mid$0.54)} & u/r & 0.71 \tiny{(0.93$\mid$0.58)} & 3/3 \\
\hline
CODI & 0.75 \tiny{(0.93$\mid$0.63)} & 0.96 \tiny{(1.00$\mid$0.93)} & n/a & n/a & 0.86 \tiny{(0.97$\mid$0.78)} & 2/2 \\
\hline
YAM++ & 0.83 \tiny{(0.99$\mid$0.72)} & 0.72 \tiny{(0.99$\mid$0.56)} & u/r & n/a & 0.78 \tiny{(0.99$\mid$0.64)} & 2/2 \\
\hline
AUTOMSv2 & 0.69 \tiny{(0.97$\mid$0.54)} & n/a & n/a & n/a & 0.69 \tiny{(0.97$\mid$0.54)} & 1/1 \\
\hline
\end{tabular}
\end{center}
\vspace{-15pt}
\end{table}
There is no best systems for all benchmarks. However, MapSSS generates the best alignments in terms of F-measure, with Aroma, WeSeE and LogMapLt as followers. We observe a high variance in the results of several systems. Outliers are, for example, a high recall for Aroma with jerm, or a poor precision for GOMMA with provenance. This might depend on inter-dependencies between matching systems and datasets, and needs additional analysis requiring a deep knowledge of the evaluated systems. Such information is, in particular, useful for developers to detect and fix problems specific to their tool.

\begin{figure}[t]
\begin{center}
\includegraphics[width=0.97\textwidth]{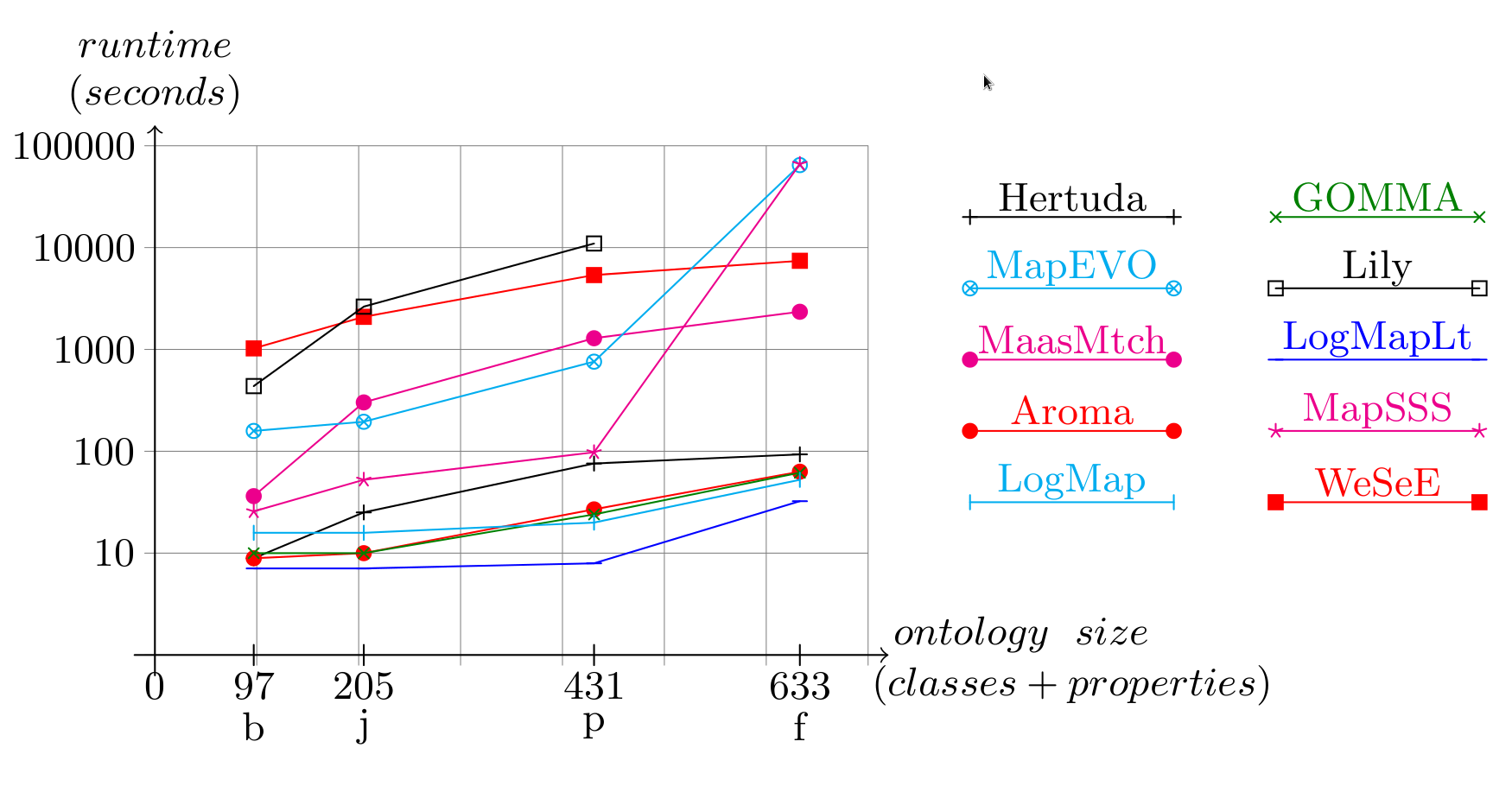}
\end{center}
\caption{Benchmark track runtimes. b=biblio, j=jerm, p=provenance, f=finance}
\label{fig:benchmarkLogGraph}
\end{figure}

Regarding runtime, a data set of 15 tests was used for each seed ontology. All  the experiments were done in a 3GHz Xeon 5472 (4 cores) 
machine running Linux Fedora 8 with 8GB RAM. Figure~\ref{fig:benchmarkLogGraph} shows a semi-log graph for runtime measurement against benchmark size in terms of classes and properties.

GOMMA, Aroma and LogMap are the fastest tools. We cannot conclude on a general correlation between runtime and quality of alignments. 
The fastest tools provide in many cases better compliance results than slower tools (MapEVO and MapPSO). However, Lily, which is the slowest tool, 
provides also alignments of high quality. Furthermore, we observe that tools are more sensitive to the number of classes and properties contained in the ontologies than to the number of 
axioms; 
the biblio and jerm ontologies have a similar number of axioms (1332 vs. 1311),
but the results for these benchmarks are different for almost all tools.

\subsection{Conference track}
\label{sec:track-conference}

For OAEI 2011.5, the available reference alignments have been refined and
harmonized. New reference alignments have been generated as a transitive
closure computed on the original reference alignments. In order to obtain a
coherent result, conflicting correspondences (i.e. those causing an
unsatisfiability) have been manually inspected and removed. 
As a result the degree of correctness and completeness of the new
reference alignment is probably slightly better than for the old one. However,
the differences are relatively restricted.

\setlength{\tabcolsep}{3pt}
\begin{table}[t]
\caption{F-measures and groups assignment within the Conference track.}
\label{table:conference}
\vspace{-15pt}
\scriptsize
\begin{center}
\begin{tabular}{|c|c|c|c|c|c|c|c|c|c|c|}
\cline{1-5} \cline{7-11}
\textbf{Group} & \textbf{System}  & \textbf{F$_{0.5}$}  & \textbf{F$_{1}$}  & \textbf{F$_{2}$}  & \ \  & \textbf{Group} & \textbf{System} & \textbf{F$_{0.5}$}  & \textbf{F$_{1}$}  & \textbf{F$_{2}$}  \\
\cline{1-5} \cline{7-11}
\hhline{|-|-|-|-|-|~|-|-|-|-|-|}

1 & YAM++  & 0.75 & 0.71 & 0.67 & & 3 & CSA & 0.49 & 0.51 & 0.54 \\
\cline{1-5} \cline{7-11}
1 & CODI   & 0.69 & 0.63 & 0.58 & & 3 &  MaasMatch & 0.53 & 0.49 & 0.45 \\
\cline{1-5} \cline{7-11}
1 & LogMap  & 0.70 & 0.61 & 0.55 & & 3 & CIDER & 0.55 & 0.49 & 0.44 \\
\cline{1-5} \cline{7-11}
1 & AgrMaker & 0.59 & 0.57 & 0.55 & & 3 & MapSSS & 0.47 & 0.46 & 0.46 \\
\cline{1-5} \cline{7-11}
1 & WeSeEMtch  & 0.61 & 0.55 & 0.49 & & 3 & Lily & 0.37 & 0.40 & 0.43 \\
\cline{1-5} \cline{7-11}
1 & Hertuda  & 0.65 & 0.55 & 0.48 & & 3 & AROMA & 0.35 & 0.38 & 0.41 \\
\hhline{|-|-|-|-|-|} \cline{7-11}
\multicolumn{2}{|c|}{\cellcolor{grayx}LogMapLt}  & \cellcolor{grayx}0.62
& \cellcolor{grayx}0.54 & \cellcolor{grayx}0.48 & & 3 & Optima & 0.26 & 0.32 & 0.42 \\
\cline{1-5} \cline{7-11}
2 & GOMMA   & 0.67 & 0.53 & 0.44 & & 4 & LDOA & 0.12 & 0.17 & 0.28 \\
\cline{1-5} \cline{7-11}
2 & AUTOMSv2 & 0.64 & 0.52 & 0.44 & & 4 & MapPSO & 0.11 & 0.06 & 0.04 \\
\hhline{|-|-|-|-|-|} \cline{7-11}
\multicolumn{2}{|c|}{\cellcolor{grayx}Baseline-E} & \cellcolor{grayx}0.64 &
\cellcolor{grayx}0.52 & \cellcolor{grayx}0.43 & & 4 & MapEVO & 0.03 & 0.02 &
0.01 \\
\cline{1-5} \cline{7-11}

\end{tabular}
\end{center}
\vspace{-15pt}
\end{table}

Table~\ref{table:conference} shows the results of all participants with regard
to the new reference alignment. F$_{0.5}$-measure, F$_{1}$-measure and
F$_{2}$-measure are computed for the threshold that provides the highest
average F$_{1}$-measure. F$_{1}$ is the harmonic mean of precision and recall
where both are equally weighted; F$_{2}$ weights recall higher than precision
and F$_{0.5}$ weights precision higher than recall.
The matchers shown in the table are ordered according to their highest average
F$_{1}$-measure. Baselines LogMapLt and Baseline-E divide matchers into four
groups. Group 1 consists of best matchers (YAM++, CODI, LogMap, AgrMaker,
WeSeEMtch and Hertuda) having better results than baseline LogMapLt in terms of
average F$_{1}$-measure. Group 2 consists of matchers that perform worse than
baseline LogMapLt in terms of average F$_{1}$-measure but still better than
Baseline-E (GOMMA, AUTOMSv2). Group 3 (CSA, MaasMtch, CIDER, MapSSS, Lily,
AROMA and Optima) contains matchers that are worse than Baseline-E but are
better (or almost the same) in terms of average F$_{2}$-measure. Finally, group
4 consists of matchers (LDOA, MapPSO and MapEVO) performing worse than
Baseline-E with regard to all F-measures.


For better comparison with previous years we also evaluated the matching systems
with the old reference alignments. The results based on the old reference
alignments are in the most of cases better by 0.03 to 0.04
points. Regarding comparison between the OAEI 2011 and OAEI 2011.5 top matchers,
YAM++ improved its results by 0.09 percentage points and remained on the top. LogMap worsened
by 0.03 percentage points while CODI provided the same results, hence CODI
and LogMap changed their position in the order according to
F$_{1}$-measure.



%

\subsection{Multifarm track}
\label{sec:track-multifarm}

In this dataset, we distinguished between two types of test cases:\footnote{We
used a subset of the whole MultiFarm dataset, omitting the ontologies Edas and Ekaw and suppressing test cases where
Russian and Chinese are involved.} (i) those test cases where two different
ontologies have been translated in different languages; and (ii) those test
cases where the same ontology has been translated in different languages.
Significant differences between results measured for (i) and (ii) can be
observed in Table~\ref{table:multifarm}. While the three systems that implement
specific multilingual techniques (WeSeE, AUTOMSv2 and YAM++ use different
translators for translating the ontologies to English) clearly generate the best
results for test cases (i), only one of these systems is among the top systems
for type (ii). This subset is dominated by the systems YAM++, CODI, and MapSSS.

\setlength{\tabcolsep}{3pt}
\begin{table}[t]
\caption{Multifarm track: results aggregated per matcher over all languages}
\label{table:multifarm}
\vspace{-15pt}
\scriptsize
\begin{center}
\begin{tabular}{|l||c|c|c|c||c|c|c|c|}
\hline
\multirow{2}{*}{\textbf{System}}& 
\multicolumn{4}{c||}{\textbf{Type (i)}} &
\multicolumn{4}{c|}{\textbf{Type (ii)}} \\
\cline{2-9}
 & \textbf{Size} & \textbf{P}  & \textbf{R} & \textbf{F} & \textbf{Size} & \textbf{P}  & \textbf{R} & \textbf{F} \\
\hline
\hline
YAM++    & 1,838  & 0.54 & 0.39 & 0.45 & 5,838  & 0.93 & 0.48 & 0.63 \\
\hline
AUTOMSv2 & 746   & 0.63 & 0.25 & 0.36 & 1,379  & 0.92 & 0.16 & 0.27 \\
\hline
WeSeE    & 4,211  & 0.24 & 0.39 & 0.29 & 5,407  & 0.76 & 0.36 & 0.49 \\
\hline
CIDER    & 737   & 0.42 & 0.12 & 0.19 & 1,090  & 0.66 & 0.06 & 0.12 \\
\hline
MapSSS   & 1,273  & 0.16 & 0.08 & 0.10 & 6,008  & 0.97 & 0.51 & 0.67 \\
\hline
LogMap   & 335   & 0.36 & 0.05 & 0.09 & 400   & 0.61 & 0.02 & 0.04 \\
\hline
CODI     & 345   & 0.34 & 0.04 & 0.08 & 7,041  & 0.83 & 0.51 & 0.63 \\
\hline
MaasMtch & 15,939 & 0.04 & 0.28 & 0.08 & 11,529 & 0.23 & 0.23 & 0.23 \\
\hline
\rowcolor[gray]{.90}
LogMapLt & 417   & 0.26 & 0.04 & 0.07 & 387   & 0.56 & 0.02 & 0.04 \\
\hline
MapPSO   & 7,991  & 0.02 & 0.06 & 0.03 & 6,325  & 0.07 & 0.04 & 0.05 \\
\hline
CSA      & 8,482  & 0.02 & 0.07 & 0.03 & 8,348  & 0.49 & 0.36 & 0.42 \\
\hline
MapEVO   & 4,731  & 0.01 & 0.01 & 0.01 & 3,560  & 0.05 & 0.01 & 0.02 \\
\hline
\end{tabular}
\end{center}
\vspace{-15pt}
\end{table}

We can observe that systems focusing on multilingual methods provide much
better results than generic matching systems.
However, the absolute results are still not very good, if compared to the top
results of the Conference dataset (0.71 F$_{1}$-measure). From all specific
multilingual methods, the techniques implemented in YAM++ generate the best
alignments in terms of F-measure (followed by AUTOMSv2 and WeSeE). It is also an
interesting outcome to see that CIDER can generate clearly the best results
compared to all other systems with non-specific multilingual systems.


Looking for the average of all systems in test cases (i) and the different pairs of languages, the best scores are for de-en (.29) and es-pt (.26) pairs. 
We cannot neglect certain language features 
in the matching process. 
The average best F-measures were indeed observed for the pairs of 
languages that have some degree of overlap in their vocabularies (de-en, fr-pt, es-pt). 
This is somehow expected, however, we could find exceptions to this behavior. In fact, 
MultiFarm requires systems exploiting more sophisticated matching strategies than label similarity and for 
many ontologies in MultiFarm it is the case. It has to be further analysed with a deep analysis of the individual pairs of ontologies. 
Furthermore, the way the MultiFarm ontologies have been translated by the different human expert may 
have an impact in the compliance of the translations according to the original ontologies.

\subsection{Anatomy track}
\label{sec:track-anatomy}

\setlength{\tabcolsep}{3pt}
\begin{table}[t]
\caption{Anatomy track: precision, recall, recall+, F-measure, and runtimes in seconds}
\label{table:anatomy}
\vspace{-15pt}
\scriptsize
\begin{center}
\begin{tabular}{|l|c|c|c|c|c|c|c|}
\hline
\textbf{System}  & \textbf{Size} & \textbf{Precision}  &  \textbf{Recall}  &  \textbf{Recall+} & \textbf{F-measure} & \textbf{Time (s)} & \textbf{Reduction}\\
\hline
\hline
 AgrMaker$_{\text{bk}}$    & 1,436  &  0.942  &  0.892  &  0.728  &  0.917 &
 1037 & 55\%  \\
\hline
 GOMMA$_{\text{bk}}$    & 1,468  &  0.927  &  0.898  &  0.736  &  0.912 & 37   &
 61\%  \\
\hline
 CODI        & 1,305  &  0.960   &  0.827  &  0.562  &  0.888 & 1177 & 98\%  \\
\hline
 LogMap      & 1,391  &  0.918  &  0.842  &  0.588  &  0.879 & 35  & 55\%  \\
\hline
 GOMMA$_{\text{nobk}}$  & 1,270  &  0.952  &  0.797  &  0.471  &  0.868 & 43   &
 53\%  \\
\hline
 MapSSS      & 1,213  &  0.934  &  0.747  &  0.337  &  0.830  & 563  & 101\%  \\
\hline
\rowcolor[gray]{.90}
 LogMapLt    & 1,155  &  0.956  &  0.728  &  0.290   &  0.827 & -   & -  \\
\hline
 Lily        & 1,370  &  0.811  &  0.733  &  0.510   &  0.770  & 657  & 80\%  \\
\hline
 Aroma       & 1,279  &  0.751  &  0.633  &  0.344  &  0.687  & 59    & 67\% \\
\hline
 CSA         & 2,472  &  0.464  &  0.757  &  0.595  &  0.576  & 5026  & 99\%  \\
\hline
 MaasMtch    & 2,738  &  0.430   &  0.777  &  0.435  &  0.554  & 68498 & 37\% 
 \\
\hline
\end{tabular}
\end{center}
\vspace{-15pt}
\end{table}


The results for the anatomy track are presented in Table~\ref{table:anatomy}.
Top results in terms of F-measure are generated by AgrMaker$_{\text{bk}}$
and GOMMA$_{\text{bk}}$. 
These systems are closely followed
by CODI, LogMap, GOMMA$_{\text{nobk}}$, and finally (with some distance)
MapSSS. Some systems could not top the LogMapLt baseline in terms of F-measure.
However, most of these systems have higher recall scores. Low F-measure values
are caused by low precision in all of these cases. This means that those systems find a large
amount of non-trivial correspondences. 

For measuring runtimes, we have executed all systems on virtual machines with
one, two, and four cores each with 8GB RAM. Runtime results shown in
Table~\ref{table:anatomy} are based on the execution of the machines with one
core. The column rightmost shows the reduction rate that was achieved when
running the tools on the four core environment, i.e. the value is computed as runtime on a 4-core environment divided by runtime using 1-core. A matcher that scales perfectly well would achieve a value of 25\%. We executed each system three times and report on average runtimes in seconds.

The fastest systems are LogMap, GOMMA (with and without the use of background
knowledge) and AROMA. The enormous variance in measured runtimes is an
interesting result. In general, there seems to be no positive correlation
between the quality of the alignment and a long runtime. The rightmost column
shows that some systems scale well and some systems can not at all exploit a
multicore environment. AgrMaker, LogMap and GOMMA reduce their runtime on a 4-core
environment up to 50\%-65\% compared to executing the system with one core. The
top system in terms of scalability is MaasMatch; we measured a reduction up to
40\%. However, we observed that running a system with 1-core vs. 
4-cores has no effect on the order of systems. Differences in runtimes are
too strong and thus the availability of additional cores does not change this
order.

\subsection{Large BioMed track}
\label{sec:track-biomed}

We evaluated the FMA-NCI matching problem using two reference
alignments based on UMLS \cite{trackBioMed2011}. The first reference alignment
contains 3,024 correspondences and represents the \emph{original} UMLS-based alignment between
FMA and NCI \cite{umlsassessment11}. This set, however, leads to a significant
number of unsatisfiable classes when integrated with FMA and NCI. The second
reference alignment addresses this problem and presents a refined set which
contains 2,898 correspondences \cite{logmap_iswc11}.
Three tasks have been considered involving different fragments of FMA
and NCI: 



\begin{description}
	\item[Task 1] consists of matching
 two (relatively small) modules of FMA and NCI.
 The FMA module contains 3,696 classes (5\% of FMA),
 while the NCI module contains 6,488 classes (10\% of NCI).

\item[Task 2] consists of matching two (relatively
large) modules of FMA and NCI.
The FMA module contains
28,861 classes (37\%  of FMA) and the NCI module contains 25,591
classes (38\% of NCI).

\item[Task 3] consists of matching the whole FMA and NCI
ontologies, which contains 78,989 and 66,724 classes, respectively.

\end{description}


\begin{table}[t]
\caption{Results for the Large BioMed track}
\label{table:largebiomed}
\vspace{-15pt}
\scriptsize
\begin{center}
\begin{tabular}{|l||c||c||c|c|c||c|c|c||c|}
\hline


\multicolumn{10}{|c|}{\textbf{Task 1}} \\\hline\hline
\multirow{2}{*}{\textbf{System}}
& \multirow{2}{*}{\textbf{Size}} 
& \multirow{2}{*}{\textbf{Unsat.}}
& \multicolumn{3}{c||}{\textbf{Refined UMLS}}
& \multicolumn{3}{c||}{\textbf{Original UMLS}}
& \multirow{2}{*}{\textbf{Time (s)}}\\
\cline{4-9}

&
& 
&
\textbf{P}  &  \textbf{R}  &  \textbf{F} &
\textbf{P}  &  \textbf{R}  &  \textbf{F} &
\textbf{}\\

\hline
\hline

GOMMA$_{\text{bk}}$  & 2,878  & 6,292  & 0.925 & \textbf{0.918}  &
\textbf{0.921} & 0.957 & \textbf{0.910} & \textbf{0.933} & 34  \\\hline

LogMap    & 2,739  & \textbf{2}      & 0.935 & 0.884  & 0.909 & 0.952  & 0.863 & 0.905 & 20  \\\hline 

GOMMA$_{\text{nobk}}$ &2,628  & 2,130  & \textbf{0.945} & 0.857  & 0.899 &
\textbf{0.973} & 0.846 & 0.905 & 27  \\\hline

\rowcolor[gray]{.90}
LogMapLt  & 2,483  & 2,104  & 0.942 & 0.807  & 0.869 & 0.969  & 0.796  & 0.874 & 10  \\\hline

Aroma 	  & 2,575  & 7,558	& 0.802 & 0.713  & 0.755 & 0.824  & 0.702  & 0.758 & 68 \\\hline 

MaasMatch &	3,696  & 9,718	& 0.580 & 0.744  & 0.652 & 0.597  &	0.730  & 0.657 &
9,437  \\\hline

CSA 	  & 3,607  & 9.590 	& 0.514 & 0.640 & 0.570  & 0.528  & 0.629  & 0.574 &
14,414 \\\hline

MapSSS 	  & 1,483  & 565    & 0.840 & 0.430 & 0.569 & 0.860  &	0.422  & 0.566 & 571

\\\hline

\multicolumn{10}{c}{}\\
\hline

\multicolumn{10}{|c|}{\textbf{Task 2}} \\\hline\hline
\multirow{2}{*}{\textbf{System}}
& \multirow{2}{*}{\textbf{Size}}
& \multirow{2}{*}{\textbf{Unsat.}} 
& \multicolumn{3}{c||}{\textbf{Refined UMLS}}
& \multicolumn{3}{c||}{\textbf{Original UMLS}}
& \multirow{2}{*}{\textbf{Time (s)}}\\
\cline{4-9}

&
& 
&
\textbf{P}  &  \textbf{R}  &  \textbf{F} &
\textbf{P}  &  \textbf{R}  &  \textbf{F}  
& \textbf{} \\

\hline
\hline

LogMap    & 2,664  & \textbf{5}  & \textbf{0.877} & 0.806  & \textbf{0.840} &
\textbf{0.887}  & 0.782 & \textbf{0.831} & 71  \\\hline

GOMMA$_{\text{bk}}$  & 2,942  & 7,304  & 0.817 & \textbf{0.830}  & 0.823 &
0.838 & \textbf{0.815} & 0.826 & 216  \\\hline

GOMMA$_{\text{nobk}}$ &2,631  & 2,127  & 0.856 & 0.777  & 0.815 &
0.873 & 0.760 & 0.813 & 160  \\\hline

\rowcolor[gray]{.90}
LogMapLt  & 3,219  & 12,682  & 0.726 & 0.807  & 0.764 & 0.748  & 0.796  & 0.771
& 26   \\\hline

CSA 	  & 3,607  & 49.831  & 0.514 & 0.640 & 0.570  & 0.528  & 0.629  & 0.574 &
14,048 \\\hline

Aroma 	  & 3,796  & 23,298	& 0.471 & 0.616 & 0.534  & 0.484 & 0.607  & 0.539  &
2,088 \\\hline

MapSSS 	  & 2,314  & 8,401   & 0.459 & 0.366 & 0.407 & 0.471  & 0.360  & 0.408 &
20,352 \\\hline

\multicolumn{10}{c}{}\\
\hline

\multicolumn{10}{|c|}{\textbf{Task 3}} \\\hline\hline
\multirow{2}{*}{\textbf{System}}
& \multirow{2}{*}{\textbf{Size}}
& \multirow{2}{*}{\textbf{Unsat.}} 
& \multicolumn{3}{c||}{\textbf{Refined UMLS}}
& \multicolumn{3}{c||}{\textbf{Original UMLS}}
& \multirow{2}{*}{\textbf{Time (s)}}\\
\cline{4-9}

&
& 
&
\textbf{P}  &  \textbf{R}  &  \textbf{F} &
\textbf{P}  &  \textbf{R}  &  \textbf{F}  
& \textbf{} \\

\hline
\hline

LogMap    & 2,658  & \textbf{9}  & \textbf{0.868} & 0.796  & \textbf{0.830} &
\textbf{0.875}  & 0.769 & 0.819 & 126  \\\hline

GOMMA$_{\text{bk}}$  & 2,983  & 17,005  & 0.806 & \textbf{0.830}  & 0.818 &
0.826 & \textbf{0.815} & \textbf{0.820} & 1,093  \\\hline

GOMMA$_{\text{nobk}}$ &2,665  & 5,238  & 0.845 & 0.777  & 0.810 &
0.862 & 0.759 & 0.807 & 960  \\\hline

\rowcolor[gray]{.90}
LogMapLt  & 3,466  & 26,429  & 0.675 & 0.807  & 0.735 & 0.695  & 0.796  & 0.742
& 57   \\\hline

CSA 	  & 3,607  & {\tiny$>$}$10^5$  & 0.514 & 0.640 & 0.570  & 0.528  & 0.629  & 0.574 &
14,068 \\\hline

Aroma 	  & 4,080  & {\tiny$>$}$10^5$	& 0.467 & 0.657 & 0.546  & 0.480 & 0.647  & 0.551  &
9,503 \\\hline

MapSSS 	  & 2,440  & 33,186   & 0.426 & 0.359 & 0.390 & 0.438  & 0.353  & 0.391
& {\tiny$>$}$10^5$ \\\hline
\end{tabular}
\end{center}
\vspace{-15pt}
\end{table}

\smallskip
We have executed all systems in a high performance server with 16 CPUs and 10
Gb. 
Table \ref{table:largebiomed} summarizes the obtained results 
where systems
has been ordered according to the F-measure against the refined reference
alignment. 
Besides precision (P), recall (R), F-measure (F) and
runtimes we have also evaluated the coherence of the alignments when reasoning together 
with the input ontologies.\footnote{We have used the OWL 2 reasoner HermiT \cite{hermit2009}}



GOMMA (with its two configurations) and 
 LogMap are a bit ahead in terms
 of F-measure with respect to Aroma, MaasMatch, CSA and MapSSS, which could not
 top the results of our base-line LogMapLt. Furthermore, MaasMatch failed to
 complete Tasks 2 and 3.
 GOMMA$_{\text{bk}}$ obtained the best results
 in terms of recall for all three tasks and the best F-measure for 
 Task 1, while LogMap provided the best results in terms
 of precision and F-measure for Tasks 2 and 3. Finally, GOMMA$_{\text{nobk}}$
 provided the most precise alignments for Task 1.
 The use of the original UMLS-based reference alignment did
not imply important variations. 
It is worth mentioning, however, that GOMMA$_{\text{bk}}$ improves its results
when comparing with the original UMLS alignment and provides the best F-measure for
Task 3.
 
As expected, efficiency decreases as the size of the input
ontologies increases.
For example,
GOMMA$_{\text{bk}}$'s F-measure decreased from $0.921$ (Task 1) to $0.818$
(Task 3). Furthermore, GOMMA$_{\text{bk}}$'s runtime also increased from 34
seconds to more than 18 minutes. CSA is an exception since (surprisingly) maintained
exactly the same results for the three tasks.

Regarding mapping coherence, only LogMap generated an `almost' clean output in
all three tasks. Although GOMMA$_{\text{nobk}}$ also provides highly
precise output correspondences, they lead to a huge amount of unsatisfiable classes.



\section{Lessons learned and future work}
\label{sec:lessons}
In the following we summarize the most important lessons learned and raise some conclusions related to future work.

\begin{description}
	\item[Multilingual coverage] Only 3 systems are able to 
deal, at a minimal level, with the multilingual labels in MultiFarm, thus 
there is plenty of room for improvements towards a multilingual semantic web. 
We could also observe a strong correlation between the ranking in Benchmark and
the ranking in MultiFarm type (ii), for non-specific multilingual systems, while
there is no (or very weak) correlation between results for tests of types (i) and (ii).
	\item[Precision and recall] It is hard to top our baselines in terms of
	F-measure. This is related to the fact that it is not easy to detect non-trivial correspondences without a (significant) loss in precision. Nevertheless, comparing OAEI 2011.5 and OAEI 2011 there is an increase in a number of high quality matchers for tracks that have not changed (Anatomy and Conference).
	\item[Computational resources] There is a high variance in runtimes between different matching algorithms. These differences cannot be counterbalanced by additionally computational power in number of cores. At the same time, we have also seen that some systems can cope with large ontologies only with large amount of RAM.
	\item[Scalability] The Benchmark results indicate that there are two families
	of systems. Those that scale well with respect to ontology size, and those
	where we find big differences in runtimes. Moreover, we have learned that the
	relevant factor is not the number of axioms, but the number of classes and
	properties.
	\item[Coherence] As shown in the Large BioMed track even highly precise alignment sets may lead to a huge number of unsatisfiable classes. Thus, the use of techniques to assess alignment coherence is critical. However, LogMap, CODI, and YAM++ are the only systems that use such techniques.\footnote{Additional results for the Conference track show that, besides LogMap, CODI and YAM generate also coherent alignments in many cases.} In future evaluations this aspect should not be neglected.
	\item[Large ontologies] Efficiency significantly decreases
	as the size of the input ontologies increases (see Benchmark and Large BioMed
	tracks). In the OAEI 2012, however, we intend to evaluate even harder problems
	such as FMA-SNOMED and SNOMED-NCI \cite{trackBioMed2011}. Although
	these matching problems will represent another significant leap in complexity, we take our
positive experiences as an indication that
matching these ontologies is still feasible.
\end{description}

\section*{Acknowledgements} 
Some of the authors are partially supported by the EU FP7 project SEALS
(IST-2009-238975). Ond\v{r}ej \v{S}v\'{a}b{-}Zamazal has been partially
supported by the CSF grant no. P202/10/0761. Ernesto Jim\'{e}nez-Ruiz was
supported by the EPSRC project LogMap. Bernardo Cuenca Grau was supported by the
Royal Society.

\bibliographystyle{abbrv}
\bibliography{ombib,seals,track-biomed,oaeitools}

\end{document}